\documentclass[letterpaper, 10 pt, conference]{ieeeconf}  

\IEEEoverridecommandlockouts                              

\usepackage{graphics} 
\usepackage{epsfig} 
\usepackage{times} 
\usepackage{amsmath} 
\usepackage{amssymb}  
\usepackage{subfigure}
\usepackage{color}
\usepackage{multirow}
\usepackage{mathtools}
\usepackage{fancyvrb}
\usepackage{hyperref}

\usepackage{subfigure}
\usepackage{booktabs}
\usepackage{bigstrut}
\usepackage{makecell}
\usepackage[table]{xcolor}
\usepackage{rotating}
\usepackage{bbding}
\usepackage[font=small]{caption}

\usepackage{soul}
\usepackage{cite}
\usepackage{dblfloatfix}

\title{\LARGE \bf
MAVIS: Multi-Camera Augmented Visual-Inertial SLAM \\using $\mathbf{SE_2(3)}$ Based Exact IMU Pre-integration}

\author{Yifu Wang$^{*}$, Yonhon Ng$^{*}$, Inkyu Sa, \'Alvaro Parra, Cristian Rodriguez-Opazo, Taojun Lin$^{\dagger}$, Hongdong Li$^{\dagger}$
\thanks{$^{*}$ equal contribution, XR Vision Labs, Tencent.}%
\thanks{$^{\dagger}$ANU, work completed while with Tencent.}
}

\begin{document}

\maketitle
\thispagestyle{empty}
\pagestyle{empty}

\begin{abstract}
We present a novel optimization-based Visual-Inertial SLAM system designed for multiple partially overlapped camera systems, named MAVIS. Our framework fully exploits the benefits of wide field-of-view from multi-camera systems, and the metric scale measurements provided by an inertial measurement unit (IMU). We introduce an improved IMU pre-integration formulation based on the exponential function of an automorphism of $\mathbf{SE_2(3)}$, which can effectively enhance tracking performance under fast rotational motion and extended integration time. Furthermore, we extend conventional front-end tracking and back-end optimization module designed for monocular or stereo setup towards multi-camera systems, and introduce implementation details that contribute to the performance of our system in challenging scenarios. The practical validity of our approach is supported by our experiments on public datasets. Our MAVIS won the first place in all the vision-IMU tracks (single and multi-session SLAM) on Hilti SLAM Challenge 2023 with 1.7 times the score compared to the second place\footnote{\scriptsize\url{https://hilti-challenge.com/leader-board-2023.html}}. 
\end{abstract}

\section*{Multimedia Material}
\noindent Supplemental Video: {\small\url{https://youtu.be/Q_jZSjhNFfg}}\\
\section{Introduction}
Robust and real-time Simultaneous Localization And Mapping (SLAM) is a long-standing problem within the computer vision and robotics communities. Pure vision-based solutions often lack the robustness and accuracy present in lidar-based solutions. Consequently, they are frequently enhanced with additional sensors, such as a low-cost IMU measuring angular velocity and acceleration, particularly on XR (VR/AR) virtual and augmented reality devices. While existing monocular or stereo visual-inertial solutions~\cite{Qin2018VINSMONO,leutenegger2015keyframe,delmerico2018benchmark,van2023eqvio,qin2019general,Geneva2020OPENVINS,rosinol2020kimera,usenko2019visual,campos2021orb} have demonstrated their potential to enhance robustness in degenerate scenarios such as texture-less environments or agile motion by integrating IMU measurements, there are still existing challenges such as limited camera field-of-view, and a restricted ability to handle feature tracking failures for long durations. These challenges can lead to rapid system divergence, even when using IMUs, causing a significant degradation in positioning accuracy.

\begin{figure}[t]
    \centering
    \includegraphics[width=0.9\columnwidth]{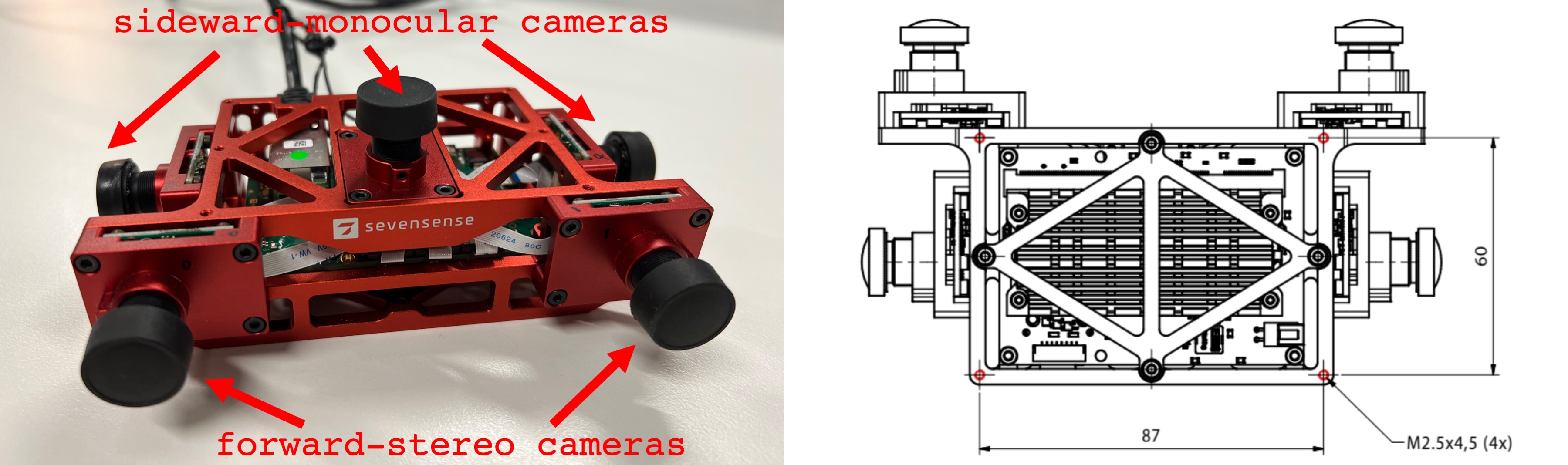}
    \caption{\label{fig:camera_model}AlphaSense multi-camera module as an example of multi-camera system analyzed in this paper, with a forward-facing stereo cameras and multiple sideward monocular cameras. }
    \vspace{-0.6cm}
\end{figure}

The present paper focuses on yet another type of sensor systems, namely multi-camera systems. As shown in Figure \ref{fig:camera_model}, the forward-facing stereo cameras offer a broader co-visibility area compared to the left, right, and upwards cameras, which have limited overlap with the forward-facing stereo pair. Such systems offer the advantage of a larger fields-of-view, omni-directional observation of the environment that improves motion estimation accuracy and robustness against failures due to texture-poor environments. However, an inherent limitation of such setups is that the introduction of additional cameras directly leads to an increase in computational cost. The proper handling of measurements from all cameras is crucial for balancing accuracy, robustness, and computational efficiency. 

Moreover, in order to improve the computational efficiency of optimization-based visual-inertial navigation methods without compromising accuracy, \cite{forster2015manifold} introduced an IMU pre-integration method, which combines hundreds of inertial measurements into a single relative motion constraint by pre-integrating measurements between selected keyframes. This formulation is vital for enhancing the effectiveness of front-end feature tracking across cameras and overall performance. However, existing methods~\cite{forster2015manifold,cui2021improved} rely on imprecise integration of position and velocity, assuming the IMU is non-rotating between measurements. This approximation can negatively impact the accuracy of pre-integrated poses, especially during fast rotational motion and extended integration times. Our contributions are as follows.
\begin{itemize}
    \item We present MAVIS, the state-of-the-art optimization-based visual-inertial SLAM framework specifically designed for multiple partially overlapped camera system. 
    \item We introduce a new IMU pre-integration formulation based on the exponential function of an automorphism of $\mathbf{SE_2(3)}$. Our approach ensures highly accurate integration of IMU data, which directly contributes to the improved tracking performance of our SLAM system.
    \item We demonstrate a substantial advantage of MAVIS in terms of robustness and accuracy through an extensive experimental evaluation. Our method attains the first place in both the vision-only single-session and multi-session tracks of the Hilti SLAM Challenge 2023.
\end{itemize}

\section{Related Work}
The advantages and challenges of monocular or stereo visual-inertial SLAM have been discussed extensively in previous frameworks~\cite{Qin2018VINSMONO,leutenegger2015keyframe,delmerico2018benchmark,van2023eqvio,qin2019general,Geneva2020OPENVINS,rosinol2020kimera,usenko2019visual,campos2021orb}. For a comprehensive survey, please refer to~\cite{Scaramuzza2016votut} and the latest research~\cite{huang2019review}. Here, we mainly focus on vision-based solutions for multi-camera systems. \cite{urban2016multicol} extended ORB-SLAM2~\cite{murORB2} to multi-camera setups, supporting various rigidly coupled multi-camera systems. \cite{kuo2020redesign} introduced an adaptive SLAM system design for arbitrary multi-camera setups, requiring no sensor-specific tuning. Several works~\cite{furgale2013toward,wang2017scale,heng18,liu2018multi,wang2020reliable} focus on utilizing a surround-view camera system, often with multiple non-overlapping monocular cameras, or specializing in motion estimation for ground vehicles. While demonstrating advantages in robustness in complex environments, these methods exhibit limited performance in highly dynamic scenarios and minor accuracy improvements in real-world experiments.

While many multi-camera visual-inertial solutions have been presented, none achieve a perfect balance among accuracy, robustness, and computational efficiency, especially in challenging scenarios. VILENS-MC~\cite{zhang2021multi} presents a multi-camera visual-inertial odometry system based on factor graph optimization. It improves tracking efficiency through cross-camera feature selection. However, it lacks a local map tracking module and loop closure optimization, leading to reduced performance in revisited locations compared to ORB-SLAM3~\cite{campos2021orb}. BAMF-SLAM~\cite{zhang2023bamfslam} introduces a multi-fisheye VI-SLAM system that relies on dense pixel-wise correspondences in a tightly-coupled semi-pose-graph bundle adjustment. This approach delivers exceptional accuracy but demands a high-end GPU for near real-time performance.

\subsection{Inertial Preintegration}
The theory of IMU preintegration was first proposed by~\cite{lupton2009inertial,lupton2012preint}. This work involves the discrete integration of the inertial measurement dynamics in a local frame of reference, such that the bias of state dynamics can be efficiently corrected at each optimization step. \cite{Forster2015IMUPO} presents a singularity-free orientation representation on $\mathbf{SO(3)}$ manifold, incorporating the IMU preintegration into optimization-based VINS, significantly improving on the stability of~\cite{lupton2012preint}. Moreover, \cite{shen2015tightly,Qin2018VINSMONO} introduced preintegration in the continuous form using quaternions, in order to overcome the discretization effect and improve the accuracy. There are also several approaches solving this problem by using analytical solution~\cite{eckenhoff2017preint,eckenhoff2018preint} or a switched linear system~\cite{Henawy2019imu,henawy2020accurate}. Another work that is closely related to ours is introduced by~\cite{barrau2020mathematical}. It extended on-manifold pre-integration of~\cite{forster2015manifold} to the Lie group $\mathbf{SE_2(3)}$. However, their method is still limited to Euler integration for position and velocity where the orientation is assumed non-rotating between IMU measurements. 

\begin{figure}[t]
    \vspace{0.1cm}
    \centering
    \includegraphics[width=0.80\columnwidth]{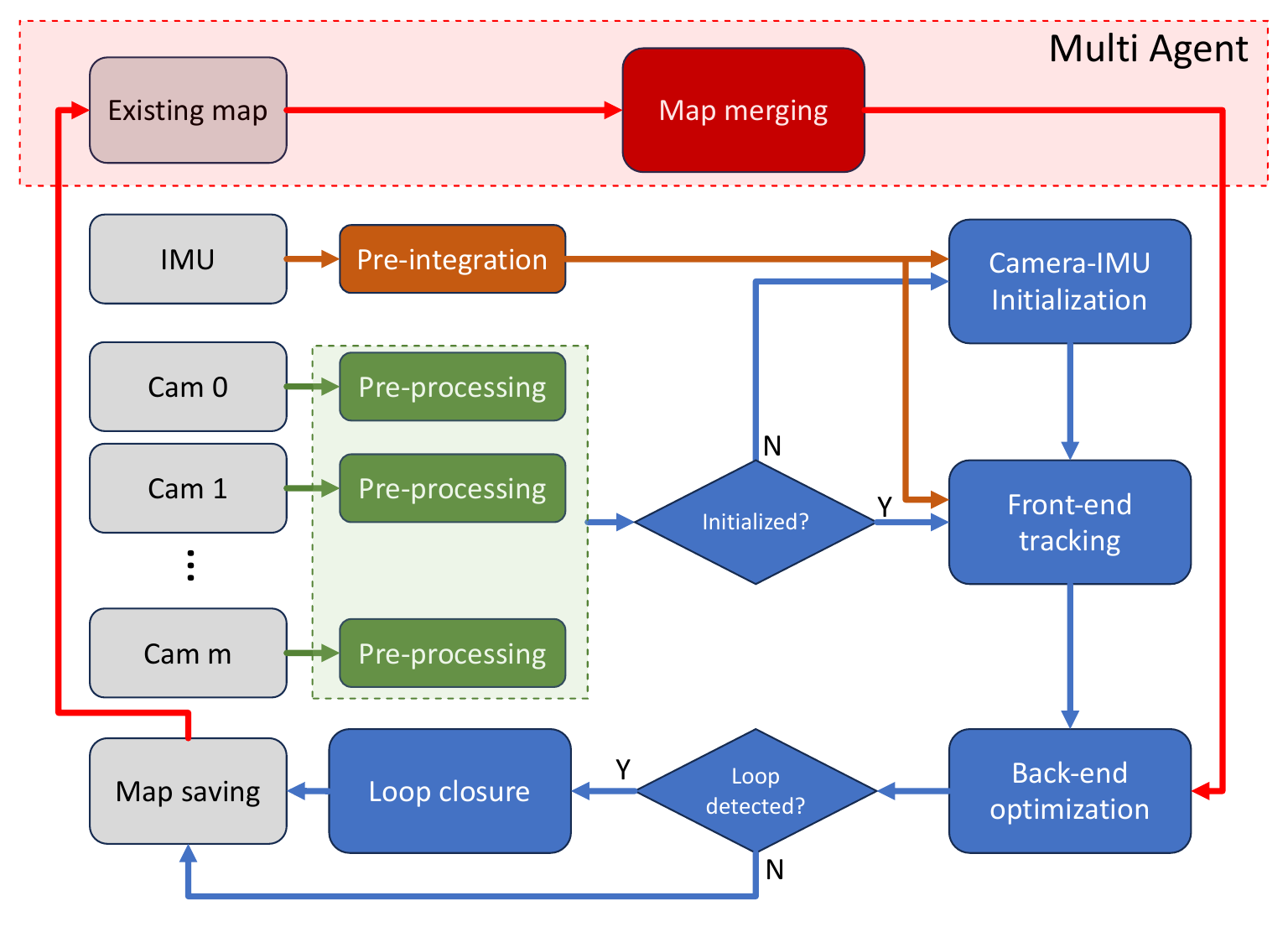}
    \caption{\label{fig:flowchart}Overview of our proposed visual-inertial localization and mapping pipeline for multi-camera systems.}
    \vspace{-0.65cm}
\end{figure}

\section{Methodology}
\label{sec:method}
In this work, we present our multi-camera VI-SLAM system with a novel automorphism of $\mathbf{SE_2(3)}$ exponential-based exact IMU pre-integration formulation, alongside an extended front-end tracking and back-end optimization modules for multi-camera setups. Figure~\ref{fig:flowchart} shows the main system components.

\subsection{IMU intrinsic compensation}
We start by adding an IMU \emph{intrinsic compensation} prior to the IMU pre-integration stage. When we use IMU information in a visual-inertial SLAM system, it is common to model raw accelerometer and gyroscope measurements as 

\vspace{-0.4cm}
\small
\begin{subequations} \label{eq:model1} 
\begin{align}
\breve{\mathbf{a}}_t &=\mathbf{a}_t+\mathbf{b}_{a_t}+\mathbf{n}_{a_t} \label{eq:model1_a}\\
\breve{\mathbf{\omega}}_t &=\mathbf{\omega}_t+\mathbf{b}_{\omega_t}+\mathbf{n}_{\omega_t},
\label{eq:model1_w}
\end{align}
\end{subequations}
\normalsize
which only considers measurement noise density (i.e., $\mathbf{n}$) and bias $\mathbf{b}$.  
Affected by acceleration bias $\mathbf{b}_{a_t}$ and gyroscope bias $\mathbf{b}_{\omega_t}$, as well as additive noises $\mathbf{n}_a$ and $\mathbf{n}_\omega$, model~\eqref{eq:model1} is simple and useful, resulting in a good approximation for devices with factory calibrated intrinsics. However, it may produce impaired calibration results for low-cost, consumer grade inertial sensors which exhibit significant axis misalignment and scale factor errors. Inspired by~\cite{rehder2016extending}, we extend the IMU model~\eqref{eq:model1} by introducing the IMU intrinsics modelling

\vspace{-0.4cm}
\small
\begin{subequations} \label{eq:model2}
\begin{align}
\breve{\mathbf{a}}_t &= \mathbf{S}_\alpha \mathbf{M}_\alpha\mathbf{a}_t+\mathbf{b}_{a_t}+\mathbf{n}_{a_t}\\
\breve{\mathbf{\omega}}_t &= \mathbf{S}_\omega \mathbf{M}_\omega\mathbf{\omega}'_t+\mathbf{A}_\omega\mathbf{a}_t+\mathbf{b}_{\omega_t}+\mathbf{n}_{\omega_t},    
\end{align}
\end{subequations}
\normalsize
whereas $\mathbf{\omega}'_t=\mathbf{C}_\omega\mathbf{\omega}_t$ is the rotation matrix between accelerometer and gyroscope. We define $\mathbf{S}_\alpha$ and $\mathbf{S}_\omega$ as the diagonal matrix of scaling effects, $\mathbf{M}_\alpha$ and $\mathbf{M}_\omega$ are lower unitriangular matrix corresponding to misalignment small angles, and let $\mathbf{A}_\omega$ be a fully populated matrix. All characteristics above can be obtained by performing the widely-adopted Kalibr~\cite{furgale2013unified}. 

\subsection{IMU pre-integration}
Following the previous section, we compensate IMU axis misalignment and scale factor, and use the notation $\bar{\omega}_t$ and $\bar{a}_t$ to denote the biased, but noise/skew-free and scale-correct IMU measurements. In the following sections, we drop or retain the subscript $t$ to simplify the notation, or highlight the time dependency of the variable. 

We now proceed to our core contribution: a novel, exact IMU pre-integration formulation based on the exponential function of an automorphism of the matrix Lie group $\mathbf{SE_2(3)}$. We first model the noise-free kinematics of the system as follow:

\vspace{-0.1cm}
\footnotesize
\begin{equation}
\left\{
  \begin{aligned}
  \dot{\mathbf{R}} &= \mathbf{R} (\bar{\mathbf{\omega}} - \mathbf{b}_\omega)^\wedge \\
  \dot{\mathbf{p}} &= \mathbf{v} \\
  \dot{\mathbf{v}} &= \mathbf{R} (\bar{\mathbf{a}} - \mathbf{b}_a) + \mathbf{g}
  \end{aligned}
\right.
\end{equation}
\normalsize
where $(\dot{\mathbf{R}}, \dot{\mathbf{p}}, \dot{\mathbf{v}})$ are the first order derivative of rotation, position and velocity of the IMU frame and $\mathbf{g}$ denotes the gravity vector, both expressed with respect to a world fixed frame. We also define the random walk of the biases $\dot{\mathbf{b}_\omega} = \mathbf{\tau}_\omega$ and $\dot{\mathbf{b}_a} = \mathbf{\tau}_a$. We can represent the IMU pose using the extended pose in $\mathbf{SE_2(3)}$ Lie group, such that the state is
\footnotesize
$ \mathbf{\xi} = \begin{bmatrix}
        \mathbf{R} & \mathbf{p} & \mathbf{v} \\
        0 & 1 & 0 \\
        0 & 0 & 1
 \end{bmatrix} 
$\normalsize
and the kinematics of the state $\mathbf{\xi}$ can be represented using an automorphism of $\mathbf{SE_2(3)}$~\cite{van2023constructive}

\vspace{-0.4cm}
\footnotesize
\begin{align}
    \label{eq:kinematics}
    \dot{\mathbf{\xi}} = (\mathbf{G} - \mathbf{D}) \mathbf{\xi} + \mathbf{\xi} (\mathbf{U} - \mathbf{B} + \mathbf{D})
\end{align}
\normalsize
where
\small
\begin{align*}
    \mathbf{U} &= \begin{bmatrix}
        \mathbf{\omega}^\wedge & 0 & \mathbf{a} \\
        0 & 0 & 0 \\
        0 & 0 & 0
    \end{bmatrix}, \hspace{0.5cm}
    \mathbf{G} = \begin{bmatrix}
        0 & 0 & \mathbf{g} \\
        0 & 0 & 0 \\
        0 & 0 & 0
    \end{bmatrix}, \\
    \mathbf{D} &= \begin{bmatrix}
        0 & 0 & 0 \\
        0 & 0 & 0 \\
        0 & 1 & 0
    \end{bmatrix}, \hspace{0.5cm}
    \mathbf{B} = \begin{bmatrix}
        (\mathbf{b}_\omega)^\wedge & 0 & \mathbf{b}_a \\
        0 & 0 & 0 \\
        0 & 0 & 0
    \end{bmatrix}
\end{align*}
\normalsize
Let us assume $t_{i-1}$ is the start time of pre-integration at previous image frame $\mathcal{F}_{i-1}$. We define $t_{j}$ be the timestamp of an arbitrary IMU measurement between frames $\mathcal{F}_{i-1}$ and $\mathcal{F}_{i}$, and let $t_{j'}$ be the timestamp of its subsequent IMU measurement, such that the small integration time $\delta = t_{j'} - t_{j}$. Let $\mathbf{\xi}_{t_{j}}$ be the extended pose at time instant $t_{j}$. 
The exact integration given~\eqref{eq:kinematics}, and assuming $\mathbf{U}$ and $\mathbf{B}$ are constant within the integration time $\delta$ is

\vspace{-0.4cm}
\small
\begin{align}
    \mathbf{\xi}_{t_{j'}} = \exp\left(\delta (\mathbf{G} - \mathbf{D})\right) \mathbf{\xi}_{t_{j}} \exp(\delta (\mathbf{U} - \mathbf{B} + \mathbf{D}))
\end{align}
\normalsize
Assuming there are $N$ sets of IMU measurements between time $t_j$ and $t_{i-1}$, we have

%
\vspace{-0.3cm}
\footnotesize
\begin{equation}
    \label{eq:integration}
    \mathbf{\xi}_{t_j} = \exp\left((\sum_{s=0}^{N} \delta_s) (\mathbf{G} - \mathbf{D})\right) \mathbf{\xi}_{t_{i-1}} \prod_{s=0}^N \exp(\delta_s (\mathbf{U}_s - \mathbf{B}_s + \mathbf{D}))
\end{equation}
\normalsize
We define $\mathcal{T} = \sum_{s=0}^{N} \delta_s$ and then rearrange~\eqref{eq:integration}, we obtain

\vspace{-0.3cm}
\footnotesize
\begin{align}
    \label{eq:multi_integration}
    \mathbf{\xi}_{t_{i-1}}^{-1} \exp\left(- \mathcal{T} (\mathbf{G} - \mathbf{D})\right) \mathbf{\xi}_{t_{j}} = \prod_{s=0}^N \exp(\delta_s (\mathbf{U}_s - \mathbf{B}_s + \mathbf{D})).
\end{align}
\normalsize
The exponential on the left equals
\footnotesize
\begin{equation}
    \label{eq:exp_g}
    \exp\left(- \mathcal{T} (\mathbf{G} - \mathbf{D})\right) = \begin{bmatrix}
        \mathbf{I} & \frac{1}{2} \mathcal{T}^2 \mathbf{g} & -\mathcal{T} \mathbf{g} \\
        0 & 1 & 0 \\
        0 & \mathcal{T} & 1
    \end{bmatrix}.
\end{equation}
\normalsize
Substituting~\eqref{eq:exp_g} into~\eqref{eq:multi_integration}, the left hand side components of~\eqref{eq:multi_integration} is exactly the same as the middle of equation (33) in~\cite{forster2015manifold}, the right hand side is our newly derived pre-integration terms, where each of the exponential can be expanded as

\vspace{-0.4cm}
\footnotesize
\begin{align}
    \label{eq:exp_preintegration}
    \exp(&\delta_s (\mathbf{U}_s - \mathbf{B}_s + \mathbf{D})) = \\
    &\begin{bmatrix}
        \exp(\delta_s (\omega_s - \mathbf{b}_{\omega_s})^\wedge) & \mathbf{J}_2 (\mathbf{a}_s - \mathbf{b}_{a_s}) & \mathbf{J}_1 (\mathbf{a}_{s} - \mathbf{b}_{a_s}) \\
        0 & 1 & 0 \\
        0 & \delta_s & 1
    \end{bmatrix},\nonumber
\end{align}
\normalsize
where
\scriptsize
\begin{align}
    \theta &= \| \tilde{\omega} \| \\
    \mathbf{J}_1 &= \delta_s \mathbf{I} + \frac{1}{\theta^2} (1 - \cos(\delta_s \theta)) \tilde{\omega}^\wedge + \frac{1}{\theta^3} (\delta_s \theta - \sin(\delta_s \theta)) (\tilde{\omega}^\wedge)^2 \label{eq:J1}\\
    \mathbf{J}_2 &= \frac{1}{2}{\delta_s}^2 \mathbf{I} + \frac{1}{\theta^3} (\delta_s \theta - \sin(\delta_s \theta)) \tilde{\omega}^\wedge + \frac{1}{\theta^4} (\frac{1}{2}{\delta_s}^2 \theta^2 + \cos(\delta_s \theta) - 1) (\tilde{\omega}^\wedge)^2 \label{eq:J2}
\end{align}
\normalsize
Here, we use $\tilde{\omega} = \omega_s - \mathbf{b}_{\omega_s}$. The iterative pre-integration is finally given as

\footnotesize
\begin{equation}
\left\{
  \begin{aligned}
   \Delta \mathbf{R}^{t_i}_{t_{j'}} &= \Delta \mathbf{R}^{t_i}_{t_j} \exp(\delta_j (\omega_j - \mathbf{b}_{\omega_j})^\wedge) \\
  \Delta \mathbf{p}^{t_i}_{t_{j'}} &= \Delta \mathbf{p}^{t_i}_{t_j} + \delta_j \Delta \mathbf{v}^{t_i}_{t_j} + \Delta \mathbf{R}^{t_i}_{t_j} \mathbf{J}_2 (\mathbf{a}_j - \mathbf{b}_{a_j}) \\
  \Delta \mathbf{v}^{t_i}_{t_{j'}} &=  \Delta \mathbf{v}^{t_i}_{t_j} + \Delta \mathbf{R}^{t_i}_{t_j} \mathbf{J}_1 (\mathbf{a}_j - \mathbf{b}_{a_j})
  \end{aligned}
\right.
\end{equation}
\normalsize
Note that under the Euler integration scheme, Jacobian terms $\mathbf{J}_1$ and $\mathbf{J}_2$ are simplified to be $\mathbf{J}_1=\delta_s \mathbf{I}$ and $\mathbf{J}_2=\frac{1}{2}{\delta_s}^2\mathbf{I}$. 

In practical implementation, the noise-free terms are not available, and are thus substituted by their corresponding estimate denoted with the $\hat{(\cdot)}$ notation. We then deal with the covariance propagation given the uncertainty of the previous estimate and measurement noise. We define the error terms as follows:

\footnotesize
\begin{equation}
    e_{ij'}=[(e^{\Delta \mathbf{R}}_{ij'})^\top,(e^{\Delta \mathbf{p}}_{ij'})^\top,(e^{\Delta \mathbf{v}}_{ij'})^\top,(e_{j'}^{\mathbf{b}_\omega})^\top,(e_{j'}^{\mathbf{b}_a})^\top]^\top
\end{equation}
\vspace{-0.5cm}
\begin{subequations}
    \begin{align}
        e^{\Delta \mathbf{R}}_{ij'} &= \log(\Delta {\mathbf{R}^{t_i}_{t_{j'}}}^\top \Delta \hat{\mathbf{R}}^{t_i}_{t_{j'}})^\vee \sim \mathcal{N}(0, \Sigma^\mathbf{R}_{ij'}) \in \mathbb{R}^3 \label{eq:e_DR} \\
        e^{\Delta \mathbf{p}}_{ij'} &= \Delta \hat{\mathbf{p}}^{t_i}_{t_{j'}} - \Delta \mathbf{p}^{t_i}_{t_{j'}} \sim \mathcal{N}(0, \Sigma^\mathbf{p}_{ij'}) \in \mathbb{R}^3 \label{eq:e_Dp}\\
        e^{\Delta \mathbf{v}}_{ij'} &= \Delta \hat{\mathbf{v}}^{t_i}_{t_{j'}} - \Delta \mathbf{v}^{t_i}_{t_{j'}} \sim \mathcal{N}(0, \Sigma^\mathbf{v}_{ij'}) \in \mathbb{R}^3 \label{eq:e_Dv}\\
        e_{j'}^{\mathbf{b}_\omega} &= \hat{\mathbf{b}}_{\omega_j} - \mathbf{b}_{\omega_j} \sim \mathcal{N}(0, \Sigma^{\mathbf{b}_\omega}_{ij'}) \in \mathbb{R}^3 \label{eq:e_bw} \\
        e_{j'}^{\mathbf{b}_a} &= \hat{\mathbf{b}}_{a_j} - \mathbf{b}_{a_j} \sim \mathcal{N}(0, \Sigma^{\mathbf{b}_a}_{ij'}) \in \mathbb{R}^3 \label{eq:e_ba}
    \end{align}
\end{subequations}
\normalsize
We derive the matrix representation of the pre-integration terms' evolution

\vspace{-0.3cm}
\footnotesize
\begin{equation}
    e_{ij'} = \mathbf{A}_{j'} e_{ij} + \mathbf{B}_{j'} \mathbf{n}_{j'} 
\end{equation}
\normalsize
where
\scriptsize
\begin{align}    
    \mathbf{A} &= \begin{bmatrix}
        \exp(- \delta_{j'} (\hat{\omega}_{j'} - \hat{\mathbf{b}}_{\omega_{j'}})^\wedge) & 0 & 0 & -\delta_{j'} \mathbf{I} & 0 \vspace{0.1cm}\\
        - \Delta \hat{\mathbf{R}}^{t_i}_{t_j}(\hat{\mathbf{J}}_2 (\hat{\mathbf{a}}_{j'} - \hat{\mathbf{b}}_{a_{j'}}))^\wedge & \mathbf{I} & \delta_{j'} \mathbf{I} & \mathbf{D}_1 & -\Delta \hat{\mathbf{R}}^{t_i}_{t_j} \hat{\mathbf{\mathbf{J}}}_2 \vspace{0.1cm}\\
        - \Delta \hat{\mathbf{R}}^{t_i}_{t_j} (\hat{\mathbf{J}}_1(\hat{\mathbf{a}}_{j'} - \hat{\mathbf{b}}_{a_{j'}}))^\wedge & 0 & \mathbf{I} & -\mathbf{D}_2 & -\Delta \hat{\mathbf{R}}^{t_i}_{t_j} \hat{\mathbf{J}}_1 \vspace{0.1cm}\\
        0 & 0 & 0 & \mathbf{I} & 0 \\
        0 & 0 & 0 & 0 & \mathbf{I}
    \end{bmatrix} \notag \\
    &\mathbf{B}=\begin{bmatrix}
        \delta_{j'} \mathbf{I} & 0 & 0 & 0 \\
        -\mathbf{D}_1 & \Delta \hat{\mathbf{R}}^{t_i}_{t_j} \hat{\mathbf{J}}_2 & 0 & 0 \\
        \mathbf{D}_2 & \Delta \hat{\mathbf{R}}^{t_i}_{t_j} \hat{\mathbf{J}}_1 & 0 & 0 \\
        0 & 0 & -\delta_{j'} \mathbf{I} & 0 \\
        0 & 0 & 0 & -\delta_{j'} \mathbf{I} 
    \end{bmatrix}, \mathbf{n}_{j'}=\begin{bmatrix}
        \mathbf{n}_{\omega_{j'}} \vspace{0.2cm} \\
        \mathbf{n}_{\mathbf{a}_{j'}} \vspace{0.2cm} \\
        \mathbf{\tau}_{\omega_{j'}} \vspace{0.2cm} \\
        \mathbf{\tau}_{a_{j'}}
    \end{bmatrix}
\end{align}
\normalsize
and
\scriptsize
\begin{align}
    \mathbf{D}_1 &= \Delta \hat{\mathbf{R}}^{t_i}_{t_j} \left( \frac{{\delta_{j'}}^2}{\| \hat{\omega}_{j'} - \hat{\mathbf{b}}_{\omega_{j'}} \|^2} \Big(((\hat{\omega}_{j'} - \hat{\mathbf{b}}_{\omega_{j'}})^\wedge (\hat{\mathbf{a}}_{j'} - \hat{\mathbf{b}}_{a_{j'}}))^\wedge \right. \nonumber\\
    &\quad+ (\hat{\omega}_{j'} - \hat{\mathbf{b}}_{\omega_{j'}})^\wedge (\hat{\mathbf{a}}_{j'} - \hat{\mathbf{b}}_{a_{j'}})^\wedge \Big) \nonumber\\ 
    &\quad \left.+ \frac{2 {\delta_{j'}}^2}{\| \hat{\omega}_{j'} - \hat{\mathbf{b}}_{\omega_{j'}} \|^4} \Big( ((\hat{\omega}_{j'} - \hat{\mathbf{b}}_{\omega_{j'}})^\wedge)^2 (\hat{\mathbf{a}} - \hat{\mathbf{b}}_{a_{j'}}) (\hat{\omega}_{j'} - \hat{\mathbf{b}}_{\omega_{j'}})^\top \Big)  \right) \\
    \mathbf{D}_2 &= \Delta \hat{\mathbf{R}}^{t_i}_{t_{j'}} \| \hat{\omega}_{j'} - \hat{\mathbf{b}}_{\omega_{j'}} \|^4 {\delta_{j'}}^2 (\hat{\omega}_{j'} - \hat{\mathbf{b}}_{\omega_{j'}})^\wedge (\hat{\mathbf{a}}_{j'} - \hat{\mathbf{b}}_{a_{j'}}) (\hat{\omega}_{j'} - \hat{\mathbf{b}}_{\omega_{j'}})^\top
\end{align}
\normalsize
where $\hat{\omega}$ and $\hat{a}$ are the biased, noisy, and intrinsic compensated IMU measurement. 
\subsection{Front-end tracking}
\label{sec:front-end}
\begin{figure}[t]
    \vspace{0.1cm}
    \centering
    \includegraphics[width=0.85\columnwidth]{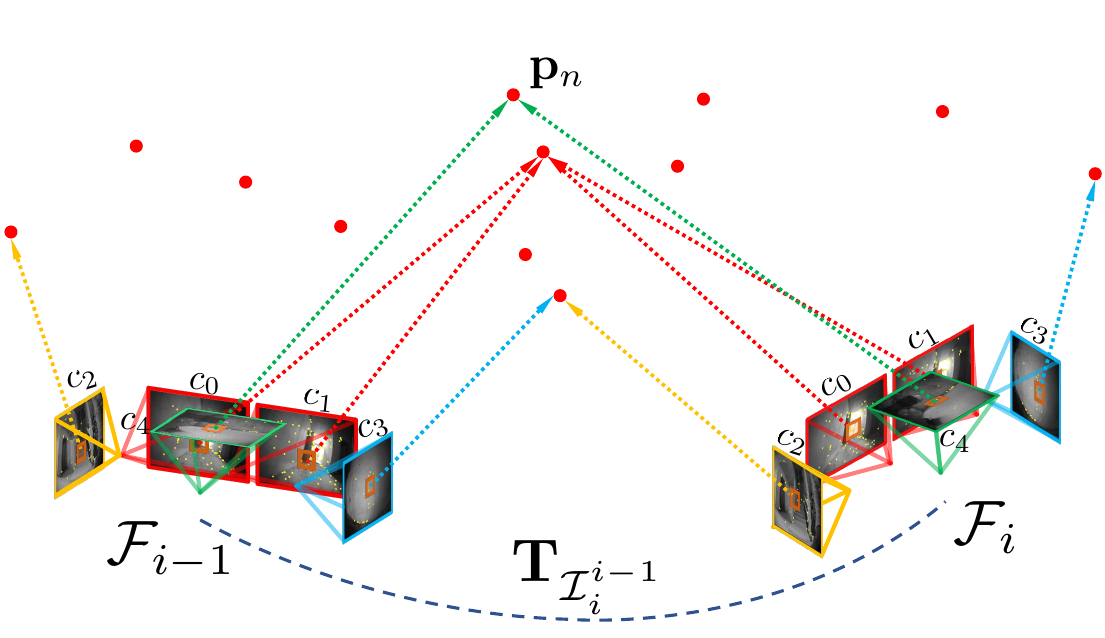}
    \caption{\label{fig:feature_matching}Geometric relationships of local feature matching across multiple cameras. }
    \vspace{-0.5cm}
\end{figure}

We aim to design a SLAM solution for sensor suites that are equipped with multiple partially overlapped cameras (cf. illustrated in Figure \ref{fig:camera_model}). Inspired by~\cite{zhang2021multi} and~\cite{campos2021orb}, we employ a localization strategy based on a local map, utilizing 2D extracted features and local 3D map points for precise pose estimation. Such local map matching mechanism can significantly improve the tracking performance when the camera revisited a previous location. We define the body frame to be the same as the IMU frame. We first project all local map points onto the multi-camera image at the current time by using the predicted relative pose obtained from IMU pre-integration. As shown in Fig \ref{fig:feature_matching}, feature matching is done for both intra-cameras and inter-cameras to enhance the co-visibility relationships. Given the multi-camera systems are precisely calibrated, the projected landmarks on an arbitrary camera can be formulated by:
\begin{equation}
(u_{c_k}^n,v_{c_k}^n) =  \pi_{c_k}(\mathbf{T}_{bc_k}^{-1}\mathbf{T}_{\mathcal{I}_i^{i-1}}^{-1}\mathbf{T}_{i-1}^{-1}\mathbf{p}_n),
\end{equation}
whereas $\mathbf{p}_n$ denotes the position of landmark $n$ in world coordinate, and $\pi_{c_k}$ be the projection function which turns $\mathbf{p}_n$ into a pixel location $(u_{c_k}^n,v_{c_k}^n)$ using intrinsic parameters of camera $c_k$. We define $\mathbf{T}_{bc_k}$ as the extrinsic parameters of camera $c_k$ with respect to the IMU coordinates. Let $\mathbf{T}_{i-1}$ be the absolute pose of reference frame $\mathcal{F}_{i-1}$ in world coordinate and $\mathbf{T}_{\mathcal{I}_i^{i-1}}$ be the estimated relative pose between current frame $\mathcal{F}_i$ and reference frame using IMU pre-integration. For those 2D features that are not associated with landmarks, we perform feature matching between the current frame and keyframes, and create new local map points through triangulation. All these relationships are then used in the back-end optimization of MAVIS to augment co-visibility edges and improve the positioning accuracy. In addition, we utilize the distance between descriptors for further validation and employ a robust cost function in the back-end optimization to eliminate all incorrectly matched feature points.

\subsection{Back-end optimization} 
Similar to many other visual-inertial SLAM/VIO systems based on optimization scheme~\cite{leutenegger2015keyframe,zhang2021multi,campos2021orb}, our MAVIS updates all the state vector by minimizing both visual reprojection errors based on observations from all cameras and error terms from pre-integrated IMU measurement using a sliding-window bundle adjustment scheme. We define the state vector $\mathcal{X}$ as the combination of motion states $\left \{ \mathbf{x} \right \} = \left [ \mathbf{x}_1 \cdots \mathbf{x}_v \right ]$ and landmarks states $\left \{ \mathbf{p} \right \} = \left [ \mathbf{p}_0 \cdots \mathbf{p}_l \right ]$, whereas $ \mathbf{x}_i$ contains 6 DoF body poses $\mathbf{R}_i$ and $\mathbf{t}_i$, linear velocity $\mathbf{v}_i$, and IMU biases $\mathbf{b}_{\omega_i}$ and $\mathbf{b}_{a_i}$. $\mathbf{p}_n$ denotes the position of the landmark $n$. The visual-inertial BA can be formulated as:
\begin{equation}\small
 \underset{\mathcal{X}}{\operatorname{min}} \sum_{i=1}^v \left(\| \mathbf{r}_{\mathbf{x}_{i},\mathcal{I}_i^{i-1}} \|^2_{\sum_{\mathbf{x}_{i},\mathcal{I}_i^{i-1}}} + \sum_{k=0}^m\sum_{n\in\mathcal{L}_i} \rho\| \mathbf{r}_{\mathbf{c}_{k},\mathbf{x}_{i},\mathbf{p}_n} \|_{\sum_{\mathbf{c}_{k},\mathbf{x}_{i},\mathbf{p}_n}} \right)
  \label{eq:windowedBA}
\end{equation}
where
\begin{itemize}
\item $\mathcal{L}_i$ is the set of landmarks observed in keyframe $\mathcal{F}_i$.
\item $\mathbf{r}_{\mathbf{x}_{i},\mathcal{I}_i^{i-1}}$ is the residual for IMU, and $\mathbf{r}_{\mathbf{c}_{k},\mathbf{x}_{i},\mathbf{p}_n}$ is the residual for visual measurements of camera $c_k$.
\item $\rho(\cdot)$ is the robust kernel used to eliminate outliers.
\end{itemize}

\section{Application to Multi-camera VI-SLAM}
An overview of our system architecture is shown in Figure~\ref{fig:flowchart}. After introducing IMU pre-integration, front-end tracking, and back-end optimization modules in Sec~\ref{sec:method}, we now proceed to the implementation particulars in this section, which directly contribute to the overall precision and robustness of our SLAM system.

\subsection{Visual measurement pre-processing}
To address the challenges in real-world application scenarios and released datasets from~\cite{zhanghilti2022}, such as dark scenes, frame drops, and data discontinuities, we conduct data pre-processing prior to the feature extraction step. Specifically, we apply histogram equalization to compensate for dark frames. 
This technique significantly enhances both the quantity and distribution of the extracted feature points. Additionally, as multi-camera devices require increased bandwidth for transmitting image data, they are more likely to encounter frame drops and synchronization issues. We therefore leverage the remaining images in feature tracking to avoid integrating IMU data for long duration. We obtain the time delay between each camera and the IMU during the camera-IMU extrinsic parameter calibration process, and add a mid-exposure time compensation parameter to further reduce the impact of synchronization errors. By adopting these approaches, we achieve visible improvements in the performance of our SLAM solution. 

\begin{table*}[!b]\scriptsize
\centering
\renewcommand\arraystretch{1.15}
\begin{tabular}{ll|ccccccccccc|c}
\hline
 &  & \textbf{MH-01} & \textbf{MH-02} & \textbf{MH-03} & \textbf{MH-04} & \textbf{MH-05} & \textbf{V-101} & \textbf{V-102} & \textbf{V-103} & \textbf{V-201} & \textbf{V-202} & \textbf{V-203} & \textbf{Std.}\\ \hline
\multicolumn{1}{l|}{\multirow{4}{*}{\textbf{\begin{tabular}[c]{@{}l@{}}Monocular\\ Inertial\end{tabular}}}} & VINS-MONO~\cite{Qin2018VINSMONO} & 0.070 & 0.050 & 0.080 & 0.120 & 0.090 & 0.040 & 0.060 & 0.110 & 0.060 & 0.060 & 0.090 & 0.025\\ 
\multicolumn{1}{l|}{} & OKVIS~\cite{leutenegger2015keyframe}& 0.160 & 0.220 & 0.240 & 0.340 & 0.470 & 0.090 & 0.200 & 0.240 & 0.130 & 0.160 & 0.290 & 0.106\\  
\multicolumn{1}{l|}{} & SVOGTSAM~\cite{delmerico2018benchmark} & 0.050 & 0.030 & 0.120 & 0.130 & 0.160 & 0.070 & 0.110 & - & 0.070 & - & - & 0.044\\  
\multicolumn{1}{l|}{} & EqVIO~\cite{van2023eqvio} & 0.176 & 0.236 & 0.112 & 0.165 & 0.238 & 0.063 & 0.128 & 0.216 & 0.058 & 0.158 & 0.176 & 0.062\\ \hline
\multicolumn{1}{l|}{\multirow{5}{*}{\textbf{\begin{tabular}[c]{@{}l@{}}Stereo\\ Inertial\end{tabular}}}} & OpenVINS~\cite{Geneva2020OPENVINS} & 0.183 & 0.129 & 0.170 & 0.172 & 0.212 & 0.055 & 0.044 & 0.069 & 0.058 & 0.045 & 0.147 & 0.063\\  
\multicolumn{1}{l|}{} & VINS-FUSION~\cite{qin2019general} & 0.181 & 0.092 & 0.167 & 0.203 & 0.416 & 0.064 & 0.270 & 0.157 & 0.065 & - & 0.160 & 0.105\\
\multicolumn{1}{l|}{} & Kimera~\cite{rosinol2020kimera} & 0.080 & 0.090 & 0.110 & 0.150 & 0.240 & 0.050 & 0.110 & 0.120 & 0.070 & 0.100 & 0.190 & 0.055\\
\multicolumn{1}{l|}{} & BASALT~\cite{usenko2019visual} & 0.080 & 0.060 & 0.050 & 0.100 & 0.080 & 0.040 & 0.020 & 0.030 & \textbf{0.030} & 0.020 & - & 0.028\\  
\multicolumn{1}{l|}{} & ORB-SLAM3~\cite{campos2020inertial} & 0.035 & 0.033 & 0.035 & \textbf{0.051} & 0.082 & 0.038 & \textbf{0.014} & 0.024 & 0.032 & \textbf{0.014} & \textbf{0.024} & 0.019\\  
\multicolumn{1}{l|}{} & \textbf{MAVIS(Ours)} & \textbf{0.024} & \textbf{0.025} & \textbf{0.032} & 0.053 & \textbf{0.075} & \textbf{0.034} & 0.016 & \textbf{0.021} & 0.031 & 0.021 & 0.039 & \textbf{0.017}\\ \hline
\end{tabular}
\caption{Performance comparison on the EuRoC~\cite{burri2016euroc} datasets, RMSE of Absolute trajectory Error (ATE) in meters.}
\label{table:euroc}
\end{table*}
\normalsize

\subsection{Camera-IMU initialization}
Inspired by~\cite{campos2021orb}, we employ a robust and accurate multi-camera IMU initialization method in our MAVIS. We firstly use stereo information from the first frame to generate an initial map. Once feature depth is known in the first keyframe, we project all landmarks in adjacent keyframes using the propagated motion model. Notably, all projections in subsequent stages are performed for both intra-camera and inter-camera scenarios. Please refer to Section \ref{sec:front-end} for detailed front-end tracking. Subsequently, we execute bundle adjustment for pure visual MAP estimation within 2 seconds while calculating IMU pre-integration and covariance between adjacent keyframes. Following the pure IMU optimization method, we jointly optimize map point positions, IMU parameters, and camera poses through multi-visual-inertial bundle adjustment. In order to avoid trajectory jumps in real-time outputs, we fix the last frame's pose during optimization. More advanced algorithms such as \cite{He_2023_CVPR} can also be employed, but is not integrated into MAVIS.

\subsection{Loop closure}

\begin{figure}[t]
    \vspace{0.1cm}
    \centering
    \includegraphics[width=0.9\columnwidth]{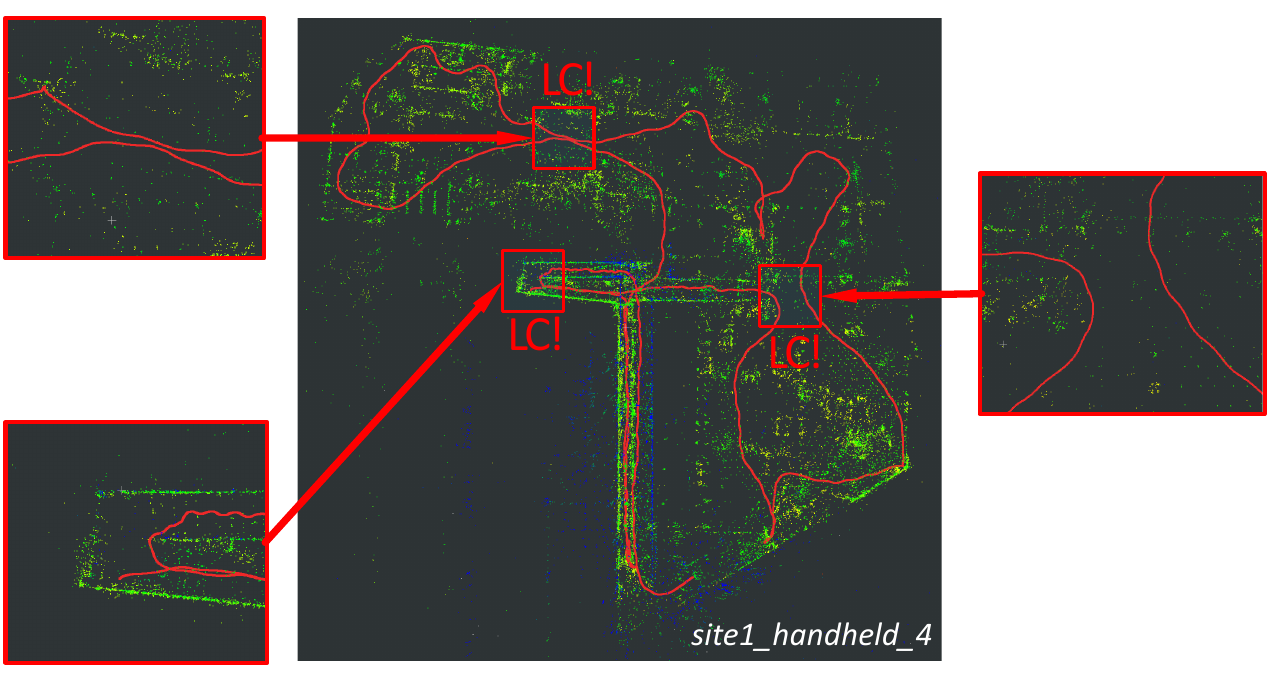}
    \caption{\label{fig:loop}Multiple detected closed loops from \textit{site1\_handheld\_4} sequence in Hilti SLAM Challenge 2023. Accumulated drifts are significantly reduced, ensuring enhanced accuracy and robustness.}
    \vspace{-0.5cm}
\end{figure}

The loop-closure module utilizes the DBoW2 library~\cite{galvez2012bags} for candidate frame detection. To fully exploit the multi-camera system's wide field-of-view, we detect putative loop closures from both intra-cameras and inter-camera, which enables our MAVIS system to correctly detect loops---like u-turn motions---that regular monocular or stereo VIO systems fail to detect (cf. illustrated in Figure \ref{fig:loop}). Similar to~\cite{campos2021orb,Qin2018VINSMONO,rosinol2020kimera}, a geometric verification is performed to remove outlier loop closures. Upon detecting a correct closed loop, we execute a global bundle adjustment, optimizing the entire trajectory with all available information, substantially reducing drift across most sequences.
\subsection{Multi-Agent SLAM} 
To participate in Hilti SLAM Challenges 2023~\cite{Hilti2023} multi-session, we adopt a similar approach for merging maps via our loop-closure correction module (refer to Figure \ref{fig:results}). We employ the Bag-of-Words (BoW) model~\cite{galvez2012bags} to identify potential overlapping keyframes, and utilizing local maps to assist geometric alignment. By fusing multiple sub-maps and conducting successful verification checks, we finally generate a globally-consistent map. In the multi-session challenge context, we designate one single-session sequence as the base for the global map. Upon the completion of SLAM processing for a sequence, the map is locally saved, and pre-loaded prior to processing subsequent data sequences. With systematic processing of all sequences and map fusion, we can integrate all submaps into a global map.

\section{Experiments}
We evaluate the performance of our method on diverse datasets. We firstly compare it to most state-of-the-art visual-inertial approaches in stereo-inertial setup using the EuRoC datasets~\cite{burri2016euroc}. We furthermore evaluate our system under multi-camera setup on the Hilti SLAM Challenge 2023 Datasets~\cite{Hilti2023}, featuring challenging sequences from handheld devices and ground robots. Both qualitative and quantitative results highlight the effectiveness of our system. Our framework is implemented in C++ and evaluated on an Ubuntu 20.04 desktop, equipped with an AMD Ryzen 9 5950X 16-Core Processor.

\begin{table*}[!t]\scriptsize
\vspace{0.2cm}
\centering
\renewcommand\arraystretch{1.15}
\begin{tabular}{cccc|c|c|c}
\hline
\multicolumn{1}{c|}{\textbf{Sequence name}} & \textbf{Difficulties} & \textbf{Sequence length} & \textbf{Processing time} & \textbf{MAVIS(Ours)} & BAMF-SLAM & Maplab2.0 \\ \hline
\multicolumn{1}{c|}{site1\_handheld\_1} & dark around stairs going to Floor 1 & 204.71s & 224.59s & \textbf{32.5} & 10.0 & 15.0 \\ 
\multicolumn{1}{c|}{site1\_handheld\_2} & dark around stairs going to Floor 2 & 167.11s & 211.98s & \textbf{23.75} & 5.0 & 12.5 \\ 
\multicolumn{1}{c|}{site1\_handheld\_3} & insufficient overlap for multi-session & 170.63s & 204.09s & \textbf{22.5} & 17.5 & 5.0 \\ 
\multicolumn{1}{c|}{site1\_handheld\_4} & - & 295.42s & 364.41s & \textbf{30.0} & 5.0 & 8.33 \\ 
\multicolumn{1}{c|}{site1\_handheld\_5} & - & 159.29s & 196.86s & \textbf{26.67} & 11.67 & 13.33 \\ 
\multicolumn{1}{l|}{site1\_multi\_session} & \multicolumn{3}{l|}{} & 4.17 & - & \textbf{5.28} \\ \hline
\multicolumn{1}{c|}{site2\_robot\_1} & unsynchronised cameras, long, no loop closure & 699.31s & 531.89s & \textbf{15.71} & 6.43 & 7.86 \\ 
\multicolumn{1}{c|}{site2\_robot\_2} & unsynchronised cameras & 305.79s & 194.40s & \textbf{53.33} & 28.33 & 15.0 \\ 
\multicolumn{1}{c|}{site2\_robot\_3} & dark, insufficient overlap for multi-session & 359.00s & 187.30s & \textbf{19.0} & 13.0 & 5.0 \\ 
\multicolumn{1}{l|}{site2\_multi\_session} & \multicolumn{3}{l|}{} & \textbf{3.33} & - & 2.33 \\ \hline
\multicolumn{1}{c|}{site3\_handheld\_1} & - & 97.18s & 124.07s & \textbf{105.0} & 45.0 & 10.0 \\ 
\multicolumn{1}{c|}{site3\_handheld\_2} & dropped data & 148.13s & 182.31s & 35.0 & \textbf{56.67} & 11.67 \\ 
\multicolumn{1}{c|}{site3\_handheld\_3} & dropped data & 189.60s & 243.46s & 23.75 & \textbf{31.25} & 7.5 \\ 
\multicolumn{1}{c|}{site3\_handheld\_4} & - & 106.88s & 130.34s & \textbf{65.0} & 37.5 & 10.0 \\ 
\multicolumn{1}{l|}{site3\_multi\_session} & \multicolumn{3}{l|}{} & \textbf{19.55} & - & 7.73 \\ \hline
\multicolumn{1}{r}{\textbf{}} & \multicolumn{3}{r|}{\textbf{Total score}} & \textbf{452.21 / 27.0} & 267.35 / - & 121.19 / 15.3 \\ \hline
\end{tabular}
\caption{Difficulties, timing and score information for test sequences}
\label{tab:results}
\vspace{-0.2cm}
\end{table*}

\begin{figure*}[!t]
    \centering
    \bgroup
    \def\arraystretch{0.5}
    \setlength\tabcolsep{0.5pt}
    \begin{tabular}{ccc}
        \begin{tabular}{c} \includegraphics[width=0.60\columnwidth]{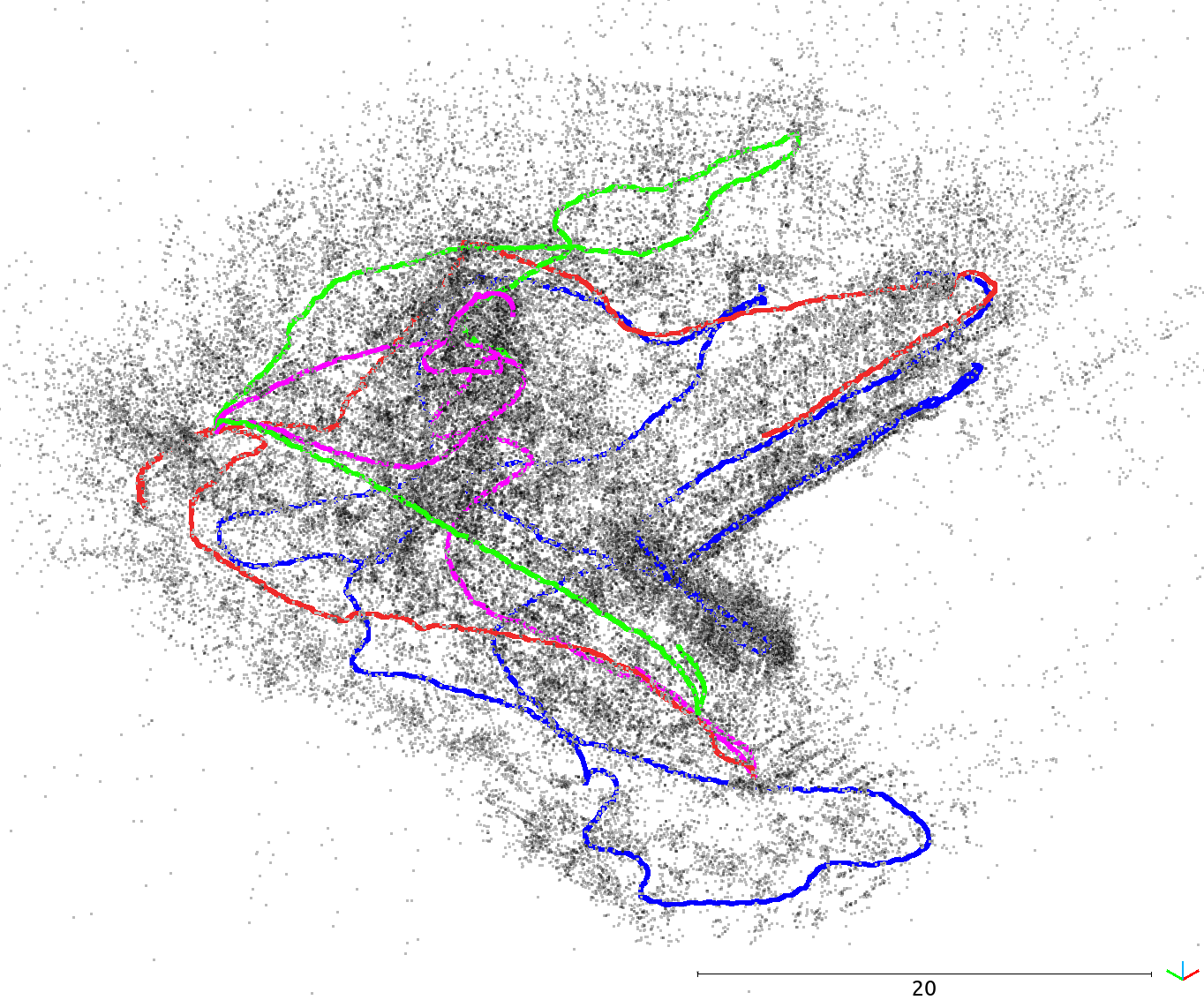} \end{tabular} & 
        \begin{tabular}{c}\includegraphics[width=0.60\columnwidth]{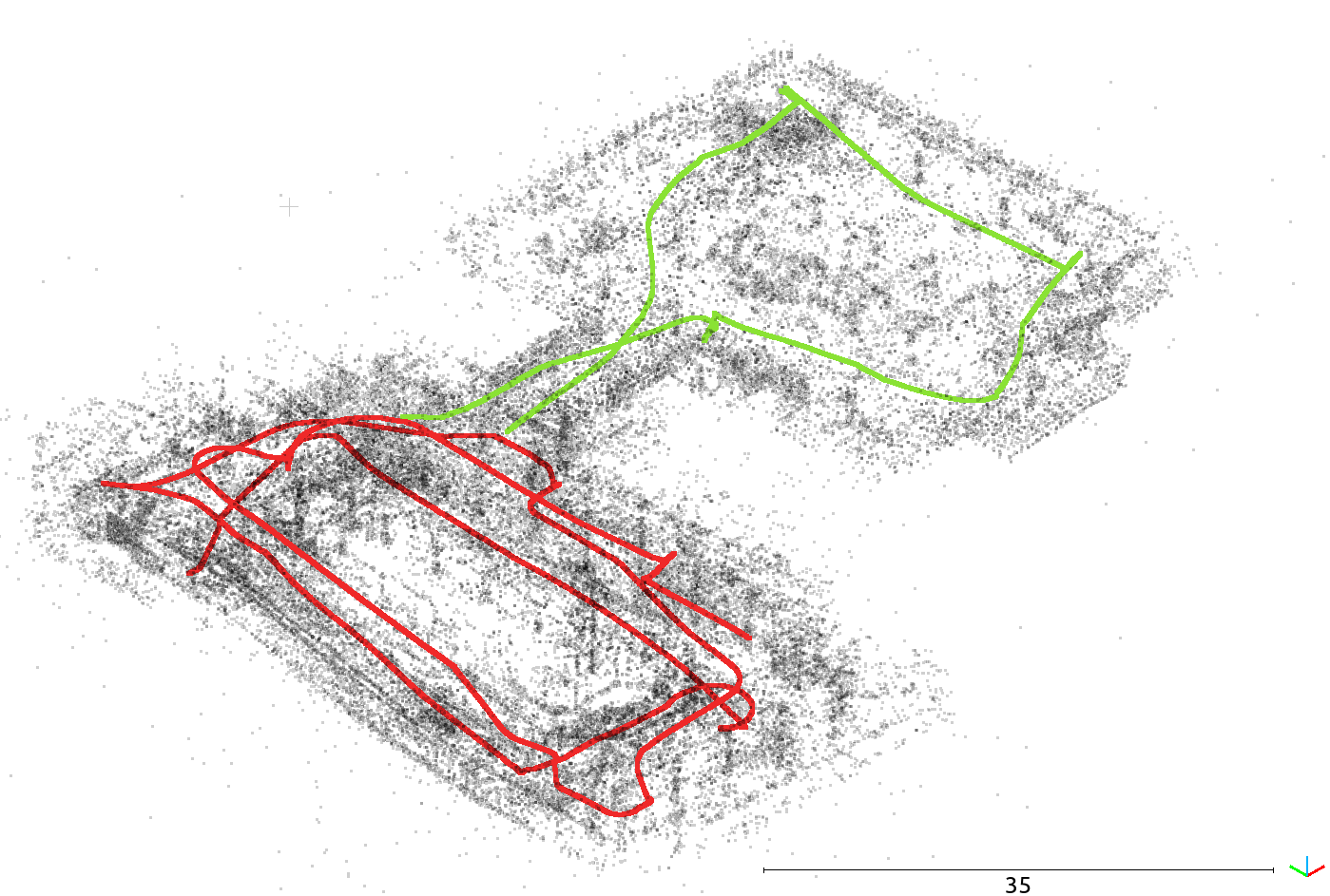}\end{tabular} & 
        \begin{tabular}{c}\includegraphics[width=0.60\columnwidth]{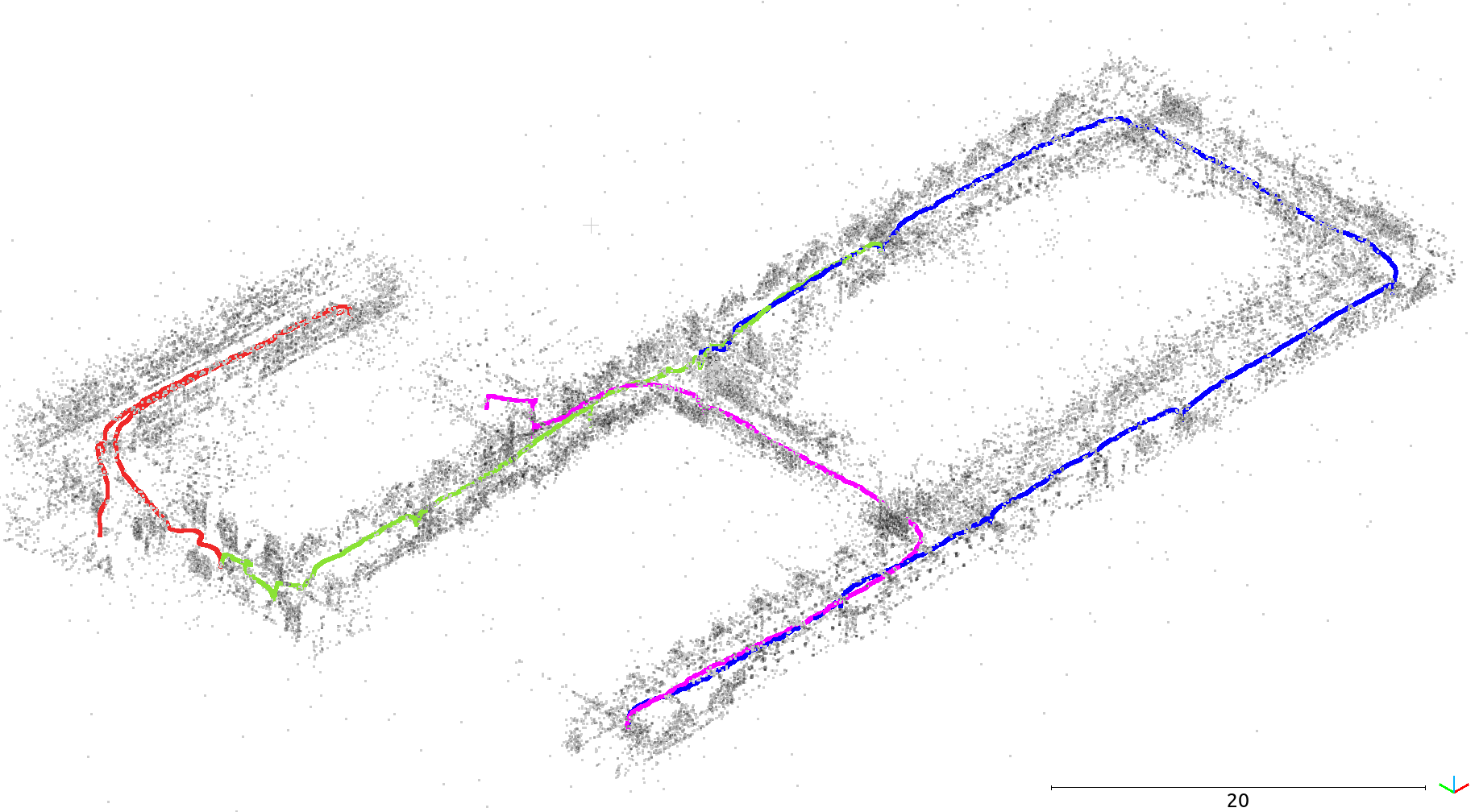}\end{tabular} \\
    \end{tabular}
    \egroup
    \caption{\label{fig:results}Perspective view of reconstructed scenes. From left to right: site\_1, site\_2, site\_3. Trajectory is coloured in sequence from red, green, blue, magenta. }
\vspace{-0.5cm}
\end{figure*}

\subsection{Performance on EuRoC datasets}
Our first experiments utilize the widely-used EuRoC datasets~\cite{burri2016euroc}, featuring sequences captured by a drone flying inside the room, equipped with synchronized stereo cameras and an IMU. We benchmark our \textbf{MAVIS} against state-of-the-art methods, including VINS-MONO~\cite{Qin2018VINSMONO}, OKVIS~\cite{leutenegger2015keyframe}, SVOGTSAM~\cite{delmerico2018benchmark}, EqVIO~\cite{van2023eqvio}, OpenVINS~\cite{Geneva2020OPENVINS}, VINS-FUSION~\cite{qin2019general}, Kimera~\cite{rosinol2020kimera}, BASALT~\cite{usenko2019visual}, and ORB-SLAM3~\cite{campos2021orb}. We employ the EuRoC dataset's calibration results and exclude our IMU intrinsic compensation for a fair comparison. We quantitatively evaluate Absolute Trajectory Error (RMSE in meters) using EVO~\cite{grupp2017evo} and summarize the results in Table~\ref{table:euroc}. Best results are in \textbf{bold}, while ``-'' indicates a method's failure to complete the sequence.

As shown in the table above, our method outperforms in most sequences. Among 11 sequences in ATE evaluations, our approach achieves the best results in 6 of them, with only ORB-SLAM3~\cite{campos2021orb} and BASALT~\cite{usenko2019visual} approaching our method. We also provide the standard deviation (std) of RMSE for all sequences, for measuring the robustness of different algorithms on the same datasets. Our method again surpasses all alternatives with a 0.017 std error. To summarize, our stereo-inertial setup demonstrates state-of-the-art performance in terms of accuracy and robustness. This could be attributed to our improved IMU pre-integration formulation, which provides more precise motion modeling and robustness in handling rapid rotations and extended integration times.

\subsection{Performance on Hilti SLAM Challenge 2023}
To thoroughly evaluate the robustness and accuracy of our multi-camera VI-SLAM system in challenging conditions, we conducted experiments on the Hilti SLAM Challenge 2023 datasets~\cite{Hilti2023}. This datasets involve handheld sequences using an Alphasense multi-camera development kit from Sevensense Robotics AG, which synchronizes an IMU with four grayscale fisheye cameras for data collection. For robot sequences, it uses four stereo OAK-D cameras and a high-end Xsens MTi-670 IMU mounted on a ground robot. We maintain consistent parameters within each dataset for our experiments. However, for the robot sequences in the Hilti Challenge 2023 datasets, we encountered inter-stereo-pair time synchronization issues in \textit{site2\_robot\_1}. Consequently, we used only a pair of stereo cameras with IMU for this sequence. For the other two robot sequences, we selected the best-synchronized four cameras in each sequence. The datasets provide millimeter-accurate ground truth on multiple control points for ATE evaluation. We compared our approach to BAMF-SLAM~\cite{zhang2023bamfslam}, ranked 2nd in the single-session vision/IMU-only track, and Maplab2.0~\cite{cramariuc2022maplab}, ranked 2nd in the multi-session track. Results are illustrated in Table~\ref{tab:results}, the following are worth noting:
\begin{itemize}
    \item We provide the difficulties, timing and score information for each sequence in Table~\ref{tab:results}. The datasets suffer from challenges such as low-light, textureless environment, unsynchronized cameras, data loss, and the absence of closed-loop exploration scenarios. While all methods successfully process the entire datasets without any gross errors.
    \item Our method achieves superior performace, clearly outperforms in both single-session and multi-session. We achieve close to 2 times the score compared to the 2nd place. More detailed analysis and further quantitative results can be found on our technical report in~\cite{Hilti2023}.
    \item Our system's runtime performance also demonstrates its practical potential. It runs in real-time on a standard desktop using only CPU. However, BAMF-SLAM~\cite{zhang2023bamfslam} requires a Nvidia GeForce RTX 4090 GPU for processing and still runs 1.6 times slower than ours. 
\end{itemize}

We also test our method in the Hilti SLAM Challenge 2022~\cite{zhanghilti2022}, achieving best results with a score of 130.2, which is three times higher than the 2nd place's score of 40.9. Please refer to the live leaderboard on~\cite{Hilti2023} for more details.

\section{Conclusion}
In this paper, we introduce MAVIS, an optimization-based visual-inertial SLAM system for multi-camera systems. Compared to alternatives, we present an exact IMU pre-integration formulation based on the $\mathbf{SE_2(3)}$ exponential, effectively improving tracking performance, especially during rapid rotations and extended integration times. We also extend front-end tracking and back-end optimization modules for multi-camera systems and introduce implementation particulars to enhance system performance in challenging scenarios. Extensive experiments across multiple datasets validate the superior performance of our method. We believe this robust and versatile SLAM system holds significant practical value for the community.

{\small
\bibliographystyle{IEEEtran}
\bibliography{egbib}
}

\end{document}